\documentclass[sigconf]{acmart}
\usepackage{multirow}
\usepackage{lscape}
\usepackage{graphicx}
\usepackage{caption}
\usepackage{wrapfig}
\usepackage{array}
\usepackage{tabularray}
\usepackage{subcaption} % For subfigures
\usepackage[linesnumbered,ruled,vlined]{algorithm2e}
\usepackage{algpseudocode}
\usepackage{hyperref}
\usepackage{comment}
\usepackage[utf8]{inputenc}
\usepackage[T1]{fontenc}

\usepackage{natbib}

\usepackage{amsmath}
\usepackage{amsmath,graphicx}

\let\algorithm\relax
\let\algorithm*\relax

\usepackage{algorithm}  % or algorithm2e
\usepackage[T1]{fontenc}
\usepackage[utf8]{inputenc}
\copyrightyear{2024}
\acmYear{2024}
\setcopyright{rightsretained}
\acmConference[CODS-COMAD Dec '24]{8th International Conference on Data Science and Management of Data (12th ACM IKDD CODS and 30th COMAD)}{December 18--21, 2024}{Jodhpur, India}
\acmBooktitle{8th International Conference on Data Science and Management of Data (12th ACM IKDD CODS and 30th COMAD) (CODS-COMAD Dec '24), December 18--21, 2024, Jodhpur, India}
\acmPrice{}
\acmDOI{10.1145/3703323.3703722}
\acmISBN{979-8-4007-1124-4/24/12}

\AtBeginDocument{%
  }

\begin{document}

\title{Action Recognition based Industrial Safety Violation Detection}

\author{Surya N Reddy}
\affiliation{%
  \institution{Indian Institute of Technology}
  \city{Bhilai}
  \country{India}}
\email{yarrabothula@iitbhilai.ac.in}

\author{Vaibhav Kurrey}
\affiliation{%
  \institution{Indian Institute of Technology}
  \city{Bhilai}
  \country{India}}
\email{vaibhavkurrey@iitbhilai.ac.in}

\author{Mayank Nagar}
\affiliation{%
  \institution{Chennai Mathematical Institute}
  \city{Chennai}
  \country{India}}
\email{nmayank1998@gmail.com}

\author{Gagan Raj Gupta}
\affiliation{%
  \institution{Indian Institute of Technology}
  \city{Bhilai}
  \country{India}}
\email{gagan@iitbhilai.ac.in}

\renewcommand{\shortauthors}{Surya et al.}

\begin{abstract}

Proper use of personal protective equipment (PPE) can save the lives of industry workers and it is a widely used application of computer vision in the large manufacturing industries. However, most of the applications deployed generate a lot of false alarms (violations) because they tend to generalize the requirements of PPE across the industry and tasks. The key to resolving this issue is to understand the action being performed by the worker and customize the inference for the specific PPE requirements of that action. In this paper, we propose a system that employs activity recognition models to first understand the action being performed and then use object detection techniques to check for violations. This leads to a 23\% improvement in the F1-score compared to the PPE-based approach on our test dataset of 109 videos.
\end{abstract}




\maketitle

\section{Introduction}

Accidents in construction sites and industrial environments can turn fatal for the workers if they don't wear proper Personal Protective Equipment (PPE). The right usage of PPE not only saves lives but also reduces the severity of the injury. Despite regulatory requirements and safety protocols, ensuring compliance with PPE 
%...
\begin{figure}[t]
  \centering
  \begin{subfigure}[b]{0.49\linewidth}
    \includegraphics[width=\linewidth]{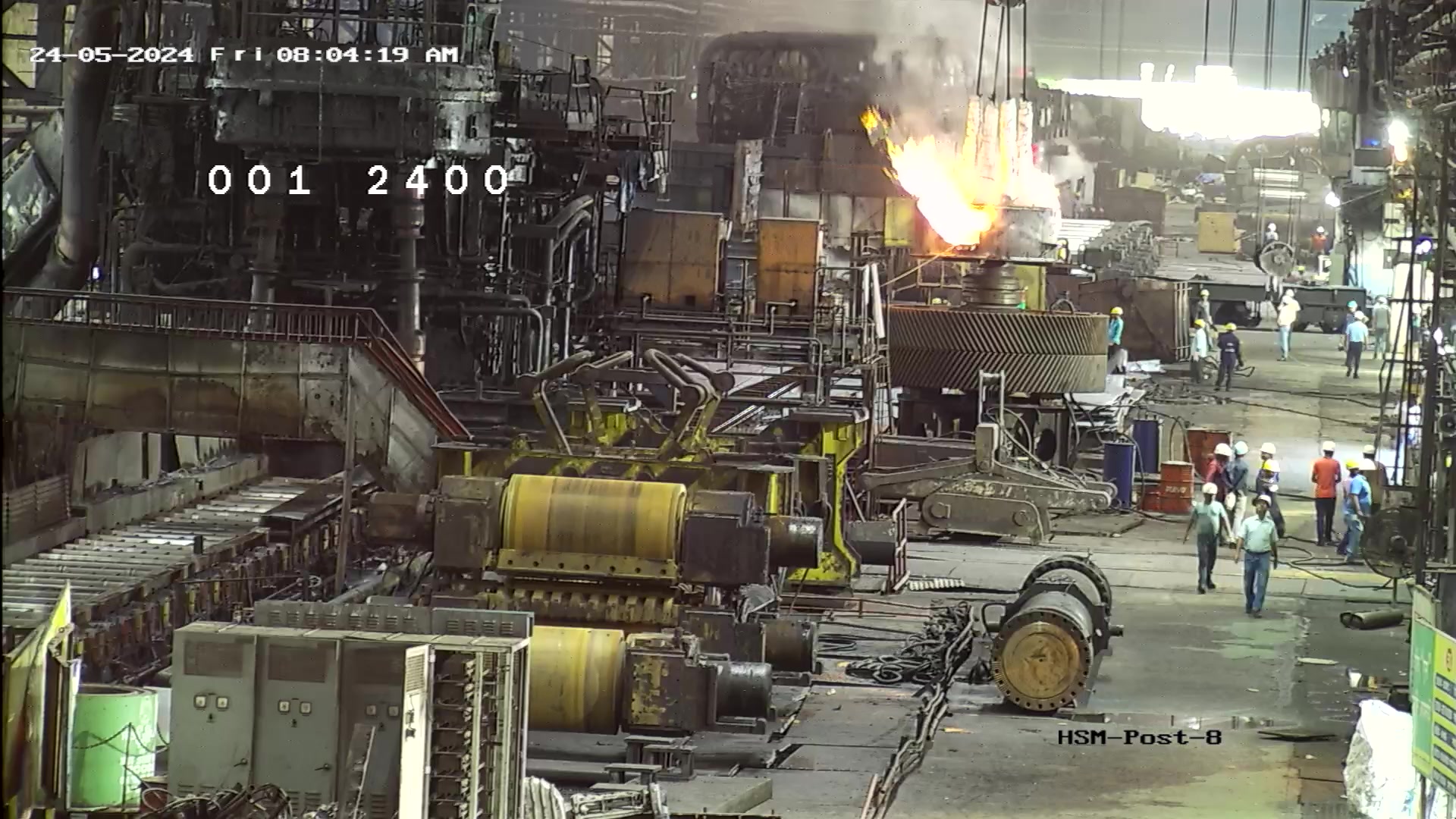}
     \caption{Real Time Surveillance Feed}
  \end{subfigure}
  \begin{subfigure}[b]{0.49\linewidth}
    \includegraphics[width=\linewidth]{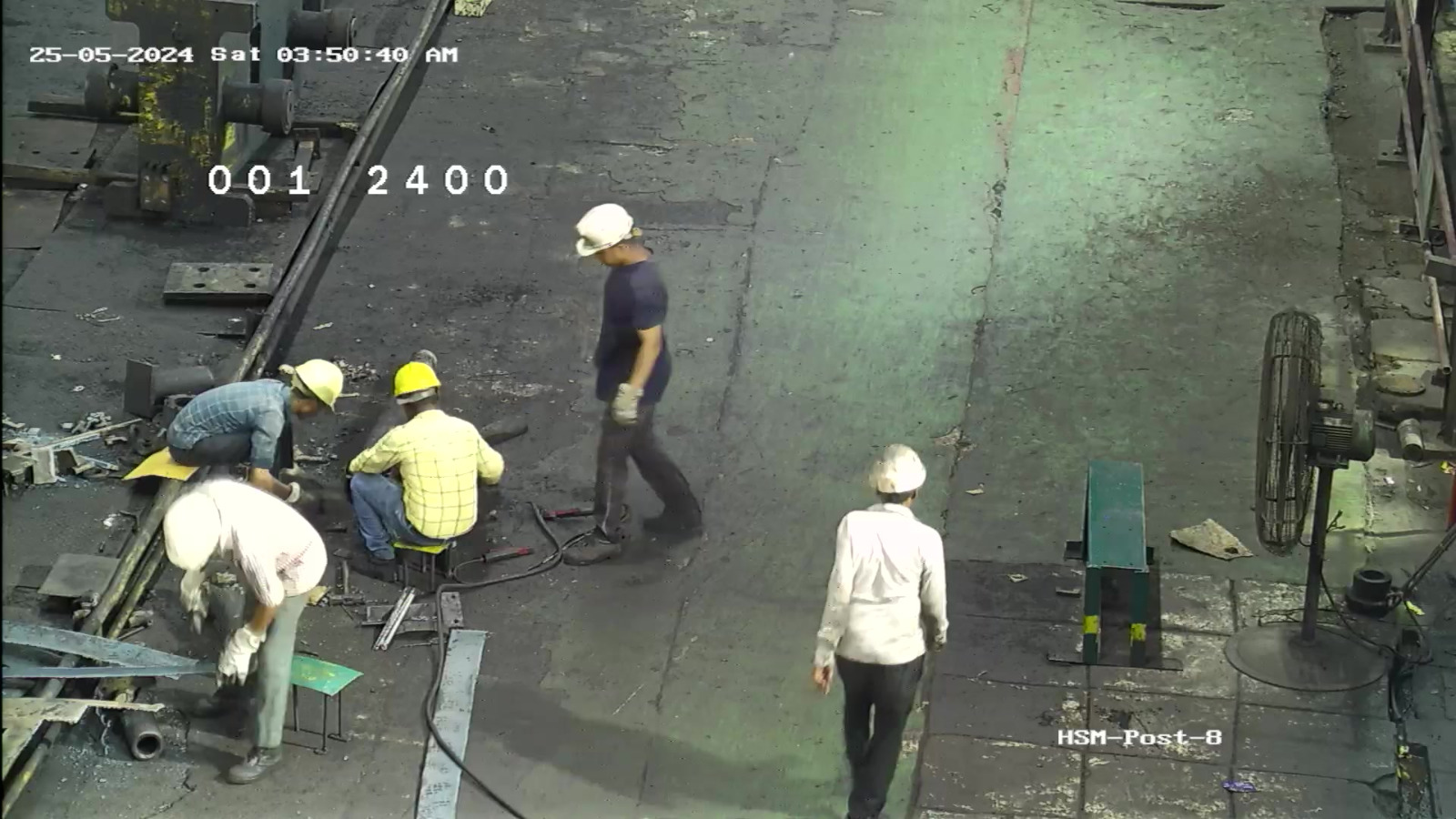}
    \caption{Multi Actions - Multi People}
  \end{subfigure}
  \begin{subfigure}[b]{0.49\linewidth}
    \includegraphics[width=\linewidth]{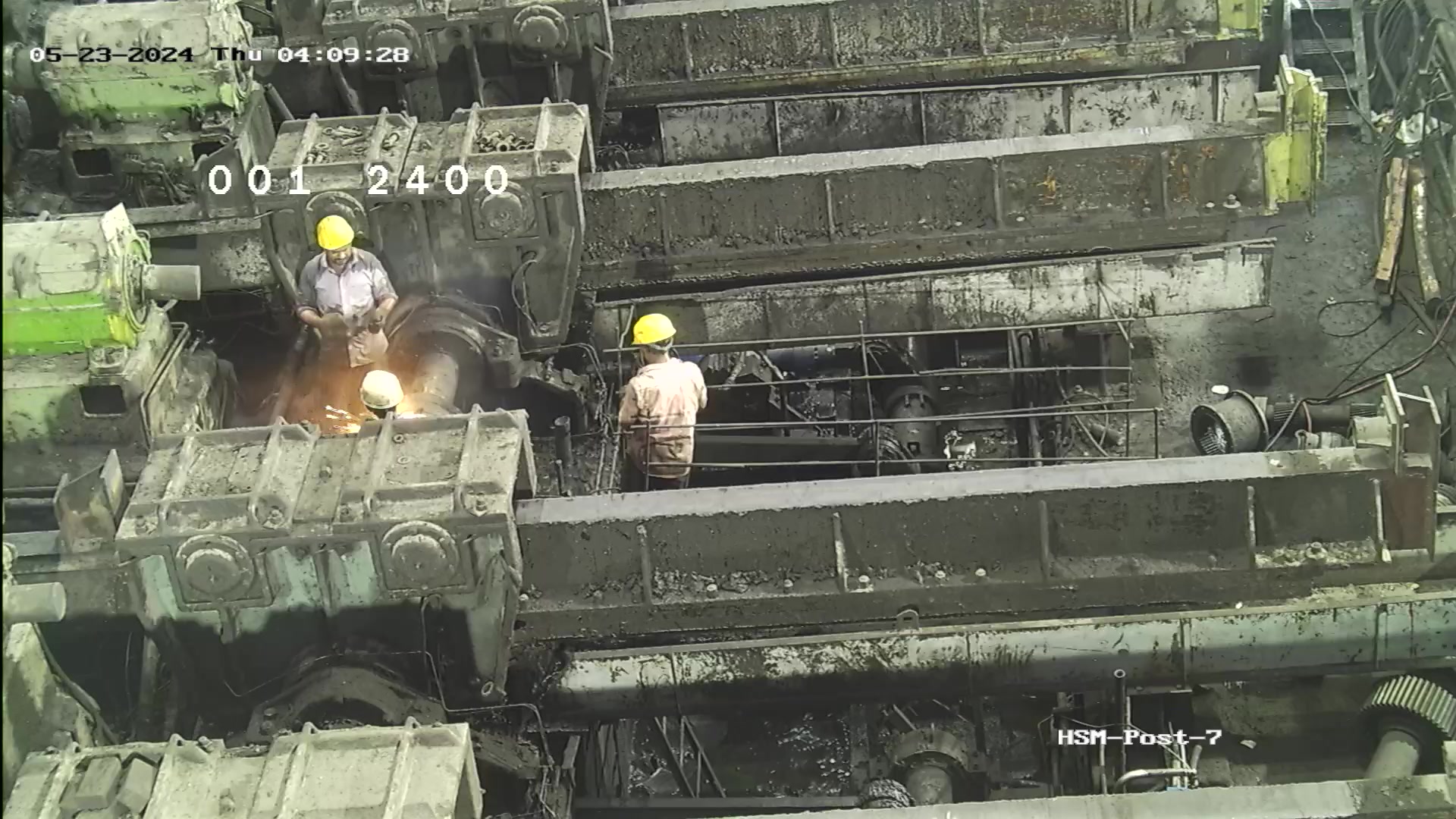}
    \caption{Welding with Occlusion}
  \end{subfigure}
  \begin{subfigure}[b]{0.49\linewidth}
    \includegraphics[width=\linewidth]{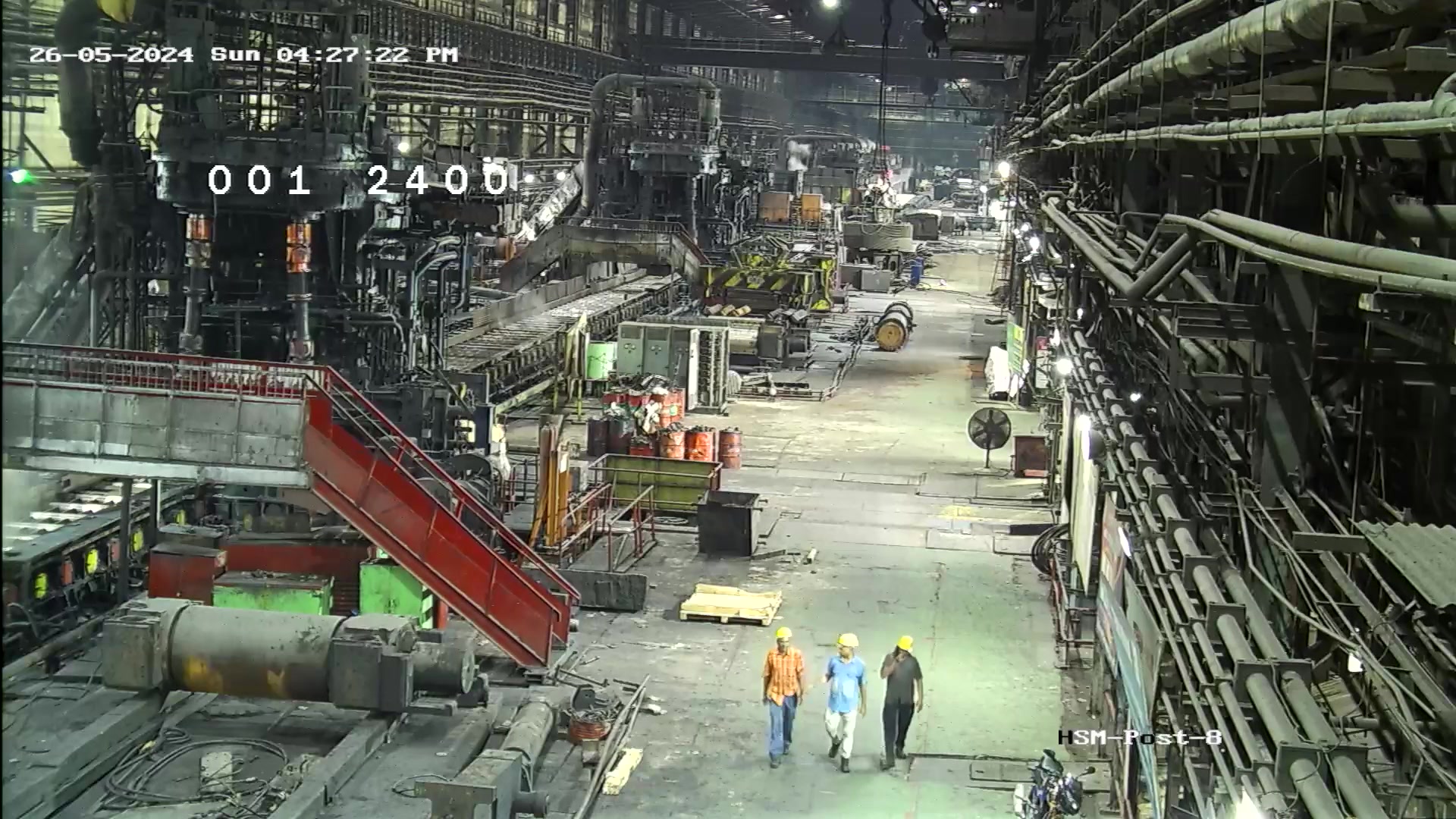}
    \caption{Multi Person Walking}
  \end{subfigure}
    
    \caption{Sample Data with Multi actor- Multi Action Industrial Scenario}
    \label{fig:main}
\end{figure}
% \vspace{-0.5cm}
requirements remains challenging for employers and safety officers. For large industries, monitoring all activities and employees is manpower-intensive and time-consuming. Video analytics-based solutions might ensure better compliance with low costs and enable the automatic detection of violations in the workplace.
During the root cause analysis of accidents, the common questions asked are: i) Was the worker wearing the PPE designated for that activity? ii) Was the worker following any unsafe practice or violating any laid down procedure during the activity? iii) Was the worker working in an unsafe environment? When multiple activities are simultaneously performed by multiple people inside a large manufacturing complex (shop floor), detecting activity-specific PPE for each person is difficult. For example, a person walking inside the shop floor might need only a Safety Helmet and Safety Shoes whereas a person working on handling materials or sharp objects also needs safety gloves. Similarly, a person doing welding needs all of the above PPEs along with safety glasses. These variations in the PPE requirements (see table~\ref{tab:ppe_requirements}) make automated violation detection difficult. Our goal in this paper is to address the questions (i) and (ii) by \textbf{building a system that can understand the activity being performed by a worker from industrial surveillance camera videos and check for any violation of PPE designated for those activities}.

Understanding human actions within the shop floor environment is crucial for developing an effective violation detection system. Sometimes, a worker may wear the necessary PPE but still engage in unsafe workflows, posing risks not only to themselves but also to surrounding workers. While action recognition and classification are well-established tasks in computer vision, there are limited published works such as InHARD \cite{9209531}, HRI30 \cite{9956300}, and LAMIS \cite{10517873} that focus on industrial actions, either in terms of model architectures or datasets. Even then, these datasets often lack the realism of an actual shop floor environment, typically featuring one person per video or focusing on a single action, failing to capture the complexities of real-world conditions (see Figure \ref{fig:main}).

Most benchmark action recognition datasets are sourced from the internet or controlled laboratory settings, predominantly featuring sports-related or household actions performed by a single actor. In real-world industrial settings, however, multiple individuals often need to be monitored simultaneously by a single camera. The existing datasets are usually well-curated, high-quality videos, which do not fully capture the dynamic and chaotic nature of real-world industrial environments. Therefore, there is a pressing need for a comprehensive dataset that authentically represents industrial actions. This paper proposes the creation of such a dataset, sourced from surveillance and process monitoring cameras within a large-scale manufacturing complex.

Integrating Human Action Recognition (HAR) models with traditional object detection systems creates a robust solution for detecting PPE violations. These models can identify specific tasks and check for compliance with PPE requirements, thereby reducing computational costs and minimizing false alarms. For effective PPE detection in real-world conditions, the dataset must be diverse, covering a wide range of tasks and PPE types. In typical industrial settings, where dedicated high-quality cameras are uncommon, the models must adapt to process monitoring or surveillance feeds, which may have poor lighting, blurry images, occlusions, and multiple individuals. These feeds often capture various activities simultaneously, making accurate violation detection challenging. Figure \ref{fig:main}
shows examples of images from real industrial settings. 

In this study, we train a SlowFast network~\cite{Feichtenhofer2018SlowFastNF} (a state-of-the-art model) for the task of video action recognition and a YOLOv9 model for PPE detection to detect industrial safety violations at a clip level. The action recognition-based PPE detection approach is compared with traditional PPE-based approaches. We also present a human study to compare the performance of our approach.

To summarize, our contributions are as follows: 
\begin{itemize}
  \item Proposed a novel dataset for understanding human actions in industrial settings.
  \item Proposed a novel approach for detecting task-specific PPE requirements using Action Recognition and Object detection models which can catch violations comparable to humans and much better than PPE-based approaches.
\end{itemize}

\begin{table*}[t]
\centering

\begin{tabular}{|c|c|c|c|c|c|}
\hline
\multicolumn{6}{|c|}{\textbf{Action Specific PPE Requirements}} \\
\hline
S.No & Activity & Shoes & Safety Helmet  & Safety Gloves & Welding Helmet \\
\hline
1 & Crane Movement & \checkmark & \checkmark &  &  \\
2 & Observing & \checkmark & \checkmark &  &  \\
3 & Interacting & \checkmark & \checkmark &  &  \\
4 & Walking & \checkmark & \checkmark &  &  \\
5 & Lifting / Handing over an object & \checkmark & \checkmark &  \checkmark &  \\
6 & Moving on Bike / Bicycle & \checkmark & \checkmark &  &  \\
7 & Pushing / Pulling an Object & \checkmark & \checkmark &   \checkmark & \\
8 & Interacting with a machine & \checkmark & \checkmark &  \checkmark & \\
9 & Doing some mechanical work & \checkmark & \checkmark &   \checkmark& \\
10 & Welding & \checkmark & \checkmark & \checkmark & \checkmark \\
\hline
\multicolumn{6}{|c|}{*Even though idle/Walking/Monitoring, PPE requirement will depend upon the workplace} \\
\hline
\end{tabular}

\caption{Action Specific PPE Requirements in Industrial Setting}
\label{tab:ppe_requirements}
\end{table*} 

\vspace{-0.27cm}
\section{Related Works}
Related works in this area can be broadly classified into three areas: video-based action recognition, publicly available datasets on HAR in an industrial context and PPE detection in industrial areas. 

\subsection{Action Recognition}
Action recognition is an area in computer vision that involves identifying and categorizing human actions in video sequences. 
Unlike static image classification, human action recognition must account for the temporal dynamics and sequential nature of actions, which significantly increases the complexity of the task \cite{Wanyan2024ACR}. 
Historically, in frame-based action recognition, there have typically been two key steps: action representation \cite{Laptev2005OnSI, Morency2007LatentDynamicDM, Scovanner2007A3S, Wang2015ARA} and action classification \cite{Kong2018HumanAR, Liu2011RecognizingHA, Shi2011HumanAS}. Recent works have merged both these approaches into an end-to-end learning framework, thus significantly improving action classification performance. 

To leverage information from all frames and model the inter-frame information correlation, Tran et al. \cite{Tran2014LearningSF} proposed 3DCNN to learn features in both spatial and temporal domains, but with high computational costs. Carreira and Zisserman \cite{Carreira2017QuoVA} introduced I3D, which builds upon the existing image classification architectures, making training easier. Feichtenhofer et al. \cite{Feichtenhofer2018SlowFastNF} proposed an efficient network, SlowFast, with both slow and fast pathways that can adapt to different scenarios by adjusting channel capacities, greatly enhancing overall efficiency. Additionally, various 3DCNN variants \cite{Feichtenhofer2020X3DEA, Tran2017ACL, Zhu2020A3DA3} have been proposed, further improving recognition efficiency and reducing the limitations in the initial architecture. ViT \cite{Dosovitskiy2020AnII}, self-attention mechanisms \cite{Devlin2019BERTPO, Vaswani2017AttentionIA} have been adapted to action recognition tasks \cite{Bertasius2021IsSA, Liu_2022}, which has been shown to achieve good performance. Spiking neural networks (SNN) have also been used for action recognition. However, due to the non-differentiability of discrete pulse signals, training SNNs poses challenges. Several effective training methods have been proposed to address this challenge \cite{leng2022differentiable, Neftci2019SurrogateGL}, but their effectiveness remains to be further investigated in the area of industrial context.

\vspace{-0.4cm}
\subsection{Action Recognition Datasets} 
Most popular action recognition datasets (see table 4), such as Weizmann \cite{Blank2005ActionsAS}, Hollywood-2 \cite{Marszalek2009ActionsIC}, HMDB \cite{Kuehne2011HMDBAL} and UCF101 \cite{Soomro2012UCF101AD}, consist of manually trimmed short clips to capture a single action. Unfortunately, these datasets don't represent real-world applications where multiple actors are working on multiple tasks and action recognition always occurs in an untrimmed environment. Video classification datasets, such as TrecVid multi-media event detection \cite{inproceedings}, Sports-1M \cite{Karpathy2014LargeScaleVC} and YouTube-8M \cite{AbuElHaija2016YouTube8MAL} have focused on video classification on a large scale by automating the label generation thereby introducing a large number of noisy annotations.

Another line of work in HAR is towards temporal localization of tasks. ActivityNet \cite{Heilbron2015ActivityNetAL}, THUMOS \cite{Idrees2016TheTC}, MultiTHUMOS \cite{Yeung2015EveryMC} and Charades \cite{Sigurdsson2016HollywoodIH} use large numbers of untrimmed videos, each containing multiple actions, obtained either from YouTube (ActivityNet, THUMOS, MultiTHUMOS) or crowd-sourced actors (Charades). The datasets cover the temporal localization aspect. However, they don't address the spatial part. 

Spatio-temporal action detection datasets, such as CMU \cite{Ke2005EfficientVE}, MSR Actions \cite{Yuan2009DiscriminativeSS}, UCF Sports \cite{Rodriguez2008ActionMA} and JHMDB \cite{Jhuang2013TowardsUA}, UCF101-24 \cite{Soomro2012UCF101AD}, AVA \cite{Gu2017AVAAV} and AVA-Kinetics \cite{Li2020TheAL}, MultiSports  \cite{Li2021MultiSportsAM} typically evaluate spatio-temporal action detection for short videos with frame-level action annotations. These benchmarks pay more attention to spatial information with frame-level detectors and clip-level detectors, which are limited to fully utilize temporal information. Very few published datasets and works are available for action recognition in the industrial context. HRI30 \cite{9956300}, InHARD \cite{9209531}, and LAMIS \cite{10517873} are some of the existing datasets available in the industrial setting. However, most of these datasets are not complex in nature and also involve one person doing a specific action (see Table \ref{tab:dataset_comparison}). In our proposed dataset we tried to capture the diversity of action and complexity of interactions between multiple actors in the video using the surveillance videos. Our dataset differs from the above in terms of both content and annotation: we label a diverse collection of industrial actions and provide spatio-temporal annotations for each subject performing an action in a large set of sampled frames.

\subsection{PPE Detection on Surveillance Videos} 
Traditional approaches to PPE detection use Object Detection (OD) models to identify the safety appliances. 
Isailovic et al. \cite{Isailovi2022TheCO} and Vukicevic et al. \cite{Vukicevic2022GenericCO} use a two-stage approach by using a key point detector to detect regions and then passing these regions to the Object Detection model for further PPE detection. 
Wu et al. \cite{Wu2019AutomaticDO} used the Single Stage Detector (SSD) \cite{Liu2015SSDSS} architecture to identify hardhats of different colors on construction sites and the model was benchmarked on GDUT-HWD \cite{Wu2019AutomaticDO} dataset. 
Otgonbold et al. \cite{Otgonbold2022SHEL5KAE} benchmarked the performance of multiple OD models for detecting 6 different classes from combinations of person, helmet, head, and face on the novel SHEL5k dataset \cite{Otgonbold2022SHEL5KAE}. 
Chen and Demachi \cite{Chen2020AVA} introduced a method using OpenPose \cite{Cao2016RealtimeM2} for body landmark detection and the YOLOv3 OD model for PPE detection. They used the geometric relationships between the key points to detect PPE and assess compliance. Zhafran et al. \cite{Zhafran2019ComputerVS} used the Fast R-CNN architecture and observed a decrease in accuracy with changes in distance and lighting conditions. 

Many existing publicly available datasets are focused on the construction industry and the manufacturing industry is largely unexplored. Most existing datasets (see Table 2) available are focused on hard hats and safety clothing. GDUT-HWD \cite{Wu2019AutomaticDO} is very noisy due to the crowd-sourced nature of data and SHW \cite{shwd24} is sourced from search engines. CHV \cite{Wang2021FastPP} dataset has no additional data and is simply a curated version of GDUT-HWD and SHW datasets. SHEL5K \cite{Otgonbold2022SHEL5KAE} and Pictor-PPE \cite{article} focus on only the PPE's clothing and hard hat aspect. These are also crowd-sourced from the web. Only TSRSF \cite{Yu2023TowardsCR} has a dataset collected from a real industrial setup in a chemical plant focusing on hard hats and clothing. However, it is close sourced. Recent work SH17 \cite{Ahmad2024SH17AD} has a comprehensive collection of PPE including gloves and earmuffs. This is again largely collected from the internet and crowd-sourced. The available datasets don't accurately reflect the environmental conditions, noise, occlusion, and lighting in the manufacturing setup and don't have a full set of PPE instances. 
Similar to SH17, we aim to bridge this gap by proposing a novel dataset collected from surveillance videos of large manufacturing industries.

% \vspace{-0.45cm}
\section{Dataset Explanation}
This paper addresses the main limitations of existing datasets in understanding industrial action i.e. they don't capture the dynamic environment and action classes that a real industrial environment presents. 
Our goal is to build a large-scale, high-quality dataset with fine-grained action classes and dense annotations that capture most of the commonly performed industrial actions. 
Also, our proposed dataset is collected from real manufacturing setup surveillance feeds and it tries to address the issue of lack of variety in the existing datasets discussed above.  

\subsection{Dataset Preparation}
\subsubsection{\textbf{Video Collection Process}}

%Video is obtained from surveillance or process monitoring cameras (2 PTZ cameras and 1 Bullet Camera) captured at a frame rate of 25 FPS, with a 1920x1080 pixels resolution. In the current version, a total of 320 hours of footage is collected and cleaned for further processing and annotations. As this footage is captured from different devices, it is first converted into a standard format as part of the data-cleaning process. This raw footage is converted to 15-second clips to understand the actions being performed in those 15 seconds. 15 seconds are chosen as this is the average time taken to understand a specific action by the human annotator. From the clips generated, we further excluded the completely noisy clips or the clips a human annotator can not understand the action being performed either due to noise or lighting or persons being distant from the camera or occlusion. From these sets, we further eliminated clips where the consecutive clips contained the same kind of action or information.   

Video is obtained from surveillance or process monitoring cameras (2 PTZ cameras and 1 Bullet Camera) captured at a frame rate of 12 FPS, with a 1920x1080 pixels resolution. In the current version, a total of 320 hours of footage is collected and cleaned for further processing and annotations.

The process involved the following key steps: 
\begin{enumerate}
\item Video Segmentation: We first divided the videos into 15-second clips to standardize the data for subsequent analysis. Human detection: Using a pre-trained person detection model, we filtered out any clips that did not contain human subjects, significantly reducing the dataset size.

\item Manual Review: The remaining clips were manually reviewed to eliminate those with poor visibility, unclear content, or other factors that made them unsuitable for future use.
\item Duplicate Removal: To ensure the uniqueness of the data set, we applied a hash-based method using the hash lib library to detect and remove duplicate clips. This step involved calculating the MD5 hash of each video file and eliminating any files with matching hashes.
\item Final Dataset: After these steps, we were left with approximately
2,900 high-quality clips, which were subsequently used for the training, testing, and validation phases of our project.

\end{enumerate}

\textbf{Clip duration:} In action recognition, 15-second video clips are chosen to provide a comprehensive temporal context for capturing and understanding actions. This duration ensures that most actions are fully represented and can be analyzed
effectively by the model. To annotate these clips, the process involves extracting 15 frames from each video. From these frames, the first 2 and last 2 are removed, leaving 11 frames for annotation. This approach helps focus on the core part of the action, reducing noise from the beginning and end of the clip, where actions may be less defined or transitional.

% \textcolor{blue}{
% \textbf{Data Annotation Process:} We adopted the AVA-style annotation and followed the step-by-step process outlined in
% this guide. Each video was cropped into 15-second clips, and one frame was extracted per second, resulting in 15 frames
% per clip. To detect humans in each frame, we used the YOLOv5 model from ultralytics YOLOv5. Following human
% detection, we utilized the VIA tool to annotate the activity being performed by the detected humans. Finally, the output
% files from the VIA tool were converted into training-ready .pkl and .csv files for use with the mmaction model.
% }

% \begin{table}[h]
%     \centering
%     \begin{tabular}{|c|c|c|c|c|}
%         \hline
%         \textbf{Dataset} & \textbf{ Classes} & \textbf{Images} & \textbf{Instances} & \textbf{Method}\\ \hline
%         Pictor - PPE\cite{article} & 3 & 784 & - & Web \\ \hline
%         SHW\cite{shwd24} & 1 & 7581 & 120558 & Web \\ \hline
%         CHV\cite{Wang2021FastPP} & 6 & 1330 & - & Web \\ \hline
%         TCRSF\cite{Yu2023TowardsCR} & 7 & 12373 & 50558 & Industrial \\ \hline
%         GDUT-HWD\cite{Wu2019AutomaticDO} & 5 & 3174 & 18893 & Web \\ \hline
%         SHEL5K\cite{Otgonbold2022SHEL5KAE} & 5 & 5000 & 75570 & Web \\ \hline
%         SH17\cite{Ahmad2024SH17AD} & 17 & 8099 & 75994 & Web \\ \hline
%         \textbf{Our Data} & \textbf{7} & \textbf{3000} & \textbf{19954} & \textbf{Industrial} \\ \hline
%     \end{tabular}
%     \caption{Existing Datasets for PPE detection}
%     \label{tab:dataset_summary}
% \end{table}

\begin{table}[h]
\centering
\begin{tabular}{|c|c|c|c|c|} 
\hline
\textbf{Dataset}                                                                             & \textbf{ Classes} & \textbf{Images} & \textbf{Instances} & \textbf{Method}      \\ 
\hline
\begin{tabular}[c]{@{}c@{}}Pictor-\\PPE \cite{article}\end{tabular}         & 3                 & 784             & -                  & Web                  \\ 
\hline
SHW \cite{shwd24}                                                           & 1                 & 7581            & 120558             & Web                  \\ 
\hline
CHV \cite{Wang2021FastPP}                                                   & 6                 & 1330            & -                  & Web                  \\ 
\hline
TCRSF \cite{Yu2023TowardsCR}                                                & 7                 & 12373           & 50558              & Industrial           \\ 
\hline
\begin{tabular}[c]{@{}c@{}}GDUT-\\HWD \cite{Wu2019AutomaticDO}\end{tabular} & 5                 & 3174            & 18893              & Web                  \\ 
\hline
SHEL5K \cite{Otgonbold2022SHEL5KAE}                                         & 5                 & 5000            & 75570              & Web                  \\ 
\hline
SH17 \cite{Ahmad2024SH17AD}                                                 & 17                & 8099            & 75994              & Web                  \\ 
\hline
\textbf{Our Data}                                                                            & \textbf{7}        & \textbf{3000}   & \textbf{19954}     & \textbf{Industrial}  \\
\hline
\end{tabular}
\caption{Existing Datasets for PPE detection}
\label{tab:dataset_summary}
\end{table}

\vspace{-0.2cm}
\subsubsection{\textbf{Action Taxonomy}}
Based on the clips collected an action taxonomy was prepared by selecting the most commonly performed actions in the videos and in the real time. Based on the inputs, the actions are fine-grained enough so that there is clarity in understanding the action and also there are not many repetitive actions. In the action dictionary provided the actions and micro-actions associated with each action are defined separately and mapping of each action with the respective classes was done. The actions defined are Crane Movement, Welding, Observing/Interacting on the Shop Floor, Walking, Moving on a Bike/Bicycle, Person Lifting/Carrying/Handing Over/Pushing or Pulling an object, Interacting with a machine/equipment on the Shop Floor. The micro action definitions provided to human annotators are shown in Table \ref{tab:class_definitions}.  

\begin{table*}[h]
    \centering
    \resizebox{\textwidth}{!}{
    \begin{tabular}{|c|c|p{15cm}|}
        \hline
        \textbf{S.No} & \textbf{Class Name} & \textbf{Definition} \\ \hline
        1 & Crane Movement & A person is performing any action when the crane is around him \\ \hline
        2 & Welding & A person is joining/separating a piece of metal with welding. Please make sure you see a welding spark while assigning this class \\ \hline
        3 & Observing & A person is simply observing an action by sitting or standing at the shop floor \\ \hline
        4 & Interacting & A person is interacting with another person by sitting or standing on the shop floor or providing instructions either non-verbally or verbally \\ \hline
        5 & Walking & A person is walking on the shop floor (Ideally without any object on his hands or head) \\ \hline
        6 & Lifting an Object & A person is lifting something with an intention to carry \\ \hline
        7 & Handing over an Object & A person is handing over an object or instrument to another person \\ \hline
        8 & Carrying an Object & A person is walking on the shop floor (Ideally with some object on his hands or head) \\ \hline
        9 & Moving on a Bike & Riding a Motorized bike on shop floor \\ \hline
        10 & Moving on a Bicycle & Riding a 2 wheeler in shop floor \\ \hline
        11 & Interacting with a machine & A person is interacting or working on a machine with someone assisting him or alone \\ \hline
        12 & Pulling or Pushing an Object & A person is pulling or pushing an object; it could be from a machine or cable \\ \hline
    \end{tabular}}
    \caption{Action Class Definitions and Descriptions provided to Human Annotators}
    \label{tab:class_definitions}

\end{table*}

\subsubsection{\textbf{Annotation Process}}

We followed the AVA \cite{Gu2017AVAAV,customava} annotation process for action labeling as this method incorporated micro-actions for understanding the entire sequence of physical activities. 
In this approach, the entire annotation process is divided into three parts person bounding box annotation, person link annotation, and action annotation. 
Person localization is done through a bounding box.
We utilized the VIA tool to annotate the activity being performed by the detected humans.

When multiple subjects are present in a selected frame, annotator evaluates the each subject separately for action annotation because action labels for each person can be different. 
Since manual bounding box annotation is intensive, a hybrid annotation approach was followed by first generating the initial set of bounding boxes using Faster-RCNN person detector \cite{Ren2015FasterRT}. 
Annotators are supplied with these proposal files to manually correct the bounding boxes generated. 
An annotator either removes incorrect bounding boxes or creates the bounding boxes missed by the person detector. 

Along the lines of annotations done for the AVA \cite{Gu2017AVAAV} dataset, bounding boxes over short periods are linked to obtaining ground-truth person tracks. The action labels (see Table 3) are generated by crowd-sourced annotators using a custom-designed interface. Each annotator will go through each frame and each person in the frame and select the corresponding micro-action for the selected person. At the bottom panel, a choice of actions is provided to the annotator for the selection. A key frame can contain multiple persons and a person can be associated with one micro-action in any given frame. Annotators have the choice not to provide any action or bounding box for the persons who are distant in the frame to reduce the noise. On average, annotators take between 60 to 180 seconds for each video depending on the number of persons present in the keyframe.  Finally, the output
files from the VIA tool were converted into training-ready files for use with the action model.

\subsection{Dataset Statistics}
\begin{figure}[htbp]
    \centering
    \includegraphics[width=0.5\textwidth]{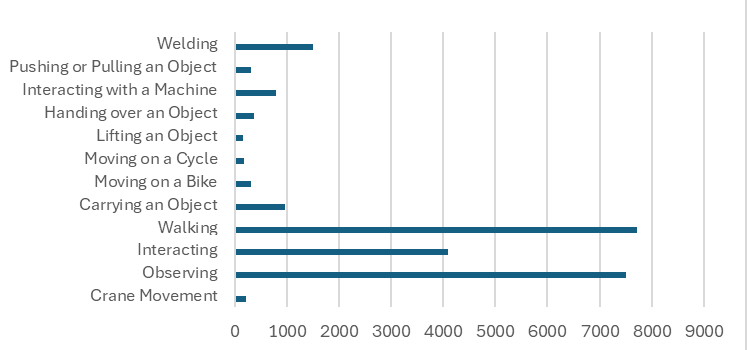}
    \caption{Distribution of Action Labels in Proposed Dataset}
    \label{fig:Action Label Distribution}
\end{figure}

Our proposed dataset is an ongoing work for building a large-scale dataset for understanding industrial actions. A total of more than 60,000 clips are generated using a total of 320 video hours. For this paper, we are presenting a total of 2900 clips annotated with micro action categories for 45652 instances distributed among 12 micro action categories. Consequently, the number of instances in our sample dataset is as high as 15.74 per video and 3511 per category. AVA-Kinetics \cite{li2020avakinetics}, which is a standard dataset for action recognition annotates only one keyframe for a 10-second clip which is much lower than ours of 10-12 keyframes per clip. As shown in Figure \ref{fig:Action Label Distribution}, the distribution of action instances is not balanced. This distribution increases the difficulty of accurately classifying the action for detection models. In our best understanding, a one-to-one comparison can not be made to any of the existing action recognition datasets as the present datasets are not oriented towards industrial actions. Also, the present action recognition databases HRI30 \cite{9956300}, LAMIS \cite{10517873}, and InHARD \cite{9209531} don't reflect the real-world setting and are very limited in scope. Even in comparison to existing non-industrial action recognition databases, our clips are longer (15 secs vs an average of 7-8 secs), with more instances per clip (15.74 vs 5 on average).

\begin{table*}[t]
    \centering
    \resizebox{\textwidth}{!}{
    \begin{tabular}{|c|c|c|c|c|c|}
        \hline
        \textbf{Dataset} & \textbf{Year} & \textbf{Data Modalities} & \textbf{Capture} & \textbf{Activities} & \textbf{Clips} \\ \hline
        MSR Action3D RGB-D Dataset \cite{Li2010ActionRB} & 2010 & Depth sequences & Depth Cameras & 20 & 320 \\ \hline
        JPL First-Person Interaction Dataset \cite{ryoo2013first} & 2013 & RGB videos & Kinect & 7 & NA \\ \hline
        HDM05 Dataset \cite{cg-2007-2} & 2007 & 3D Motion & 3D motion sensors (MoCap) & 70 & 1457 \\ \hline
        CMU Motion Capture \cite{mocap} & 2011 & 3D Motion & 3D motion sensors (MoCap) & 109 & 2605 \\ \hline
        KIT Whole-Body Human Motion \cite{7251476} & 2015 & 3D Motion, Videos & \multicolumn{1}{c|}{\begin{tabular}[c]{@{}c@{}}3D motion sensors (MoCap)\\ and a Monocular Camera\end{tabular}} & 15 & 9727 \\ \hline
        TUM Kitchen Dataset \cite{Tenorth2009TheTK} & 2009 & RGB video & \multicolumn{1}{c|}{\begin{tabular}[c]{@{}c@{}}3D motion sensors (MoCap),\\ 4 Mono Cameras, RFID Tags, Magnetic Sensors\end{tabular}} & 10 & NA \\ \hline
        HMDB51 \cite{Kuehne11} & 2014 & RGB Video & YouTube & 51 & 6766 \\ \hline
        UCF-101 \cite{soomro2012ucf101dataset101human} & 2013 & RGB Video & YouTube & 101 & 13320 \\ \hline
        MultiSports \cite{Li2021MultiSportsAM} & 2021 & RGB Video & YouTube & 66 & 3200 \\ \hline
        HRI30 \cite{Iodice2022HRI30AA} & 2022 & RGBVideo & Industrial Setting Simulated in Lab & 30 & 2940 \\ \hline
        INHARD: INdustrial Human Action Recognition Dataset \cite{Dallel2020InHARDI} & 2020 & RGB Video & Sensors, Videos & 13 & 4800 \\ \hline
        LAMIS Database \cite{10517873} & 2024 & RGB Video & Human Interactions with Lathe Machine in Workshop & 17 & 2214 \\ \hline
        \textbf{Our Dataset} & \textbf{2024} & \textbf{RGB Video} & \textbf{Industrial Surveillance Feed} & \textbf{12} & \textbf{2900} \\ \hline 

    \end{tabular}}
    \caption{Comparison of Various Action Recognition Datasets}
    \label{tab:dataset_comparison}
\end{table*}

% \vspace{-0.5cm}
\subsection{Dataset Characteristics}
One of the important goals of this work is to build a diverse and rich dataset (see Figure 2) for industrial action recognition.
Besides variation in bounding box size and size of the persons or objects in the frame, many categories will require discriminating fine-grained differences, such as “observing” versus “interacting” or lifting an object or transporting an object. 
Even within an action class, the appearance varies with vastly different contexts: an object is being simply lifted and handed over to another person or an object is being transported by two people. 
Similarly, when we detect the motion of the crane the distance of the crane from the camera is only understood through the size of the hook. 
Also, it is difficult to accurately estimate the distance between persons in the surrounding area of crane movement. 
These wide intra-class varieties will allow us to learn features that identify the critical spatio-temporal parts of action — such as in the given frame whether a person is welding or simply observing the welding process from very near.  

% \vspace{-0.5cm}
\section{Model Architecture and Metrics}
\subsection{Action Recognition}
In this study, we employ the SlowFast Networks \cite{Feichtenhofer2018SlowFastNF} for the task of video action recognition to detect industrial safety violations.
The SlowFast network is a state-of-the-art model known for its ability to capture both slow and fast visual information, making it well-suited for recognizing complex actions in videos. 
The SlowFast network has two pathways: slow and fast. The slow pathway processes video frames at a lower frame rate to capture high-resolution spatial details and static content. 
The fast pathway processes video frames at a higher frame rate to capture dynamic, rapid movements.
These pathways are fused through lateral connections, combining static and dynamic features for effective spatio-temporal pattern recognition in video data.

\subsection{PPE Detection}
To establish the baseline metric on the proposed dataset, we trained the RetinaNet \cite{lin2017focal} and Fast R-CNN models from Detectron2 \cite{wu2019detectron2}, along with the YOLOv9 \cite{wang2024yolov9} model on the dataset with 3522 images. The distribution of PPE classes is given in table \ref{tab: ppe dist}.

\textbf{RetinaNet} \cite{lin2017focal} is a one-stage object detection model that uses a focal loss function to address class imbalance during training. 
The focal loss applies a modulating term to the cross-entropy loss, focusing learning on hard negative examples. 
\textbf{Fast R-CNN} \cite{girshick2015fast} processes an entire image and a set of object proposals. 
For each object proposal, a region of interest (RoI) pooling layer extracts a fixed-length feature vector from the feature map which is fed into a sequence of fully connected layers for producing the softmax probabilities and encoding refined bounding-box positions for each class. 
\textbf{YOLOv9} \cite{wang2024yolov9} combines two neural network architectures, CSPNet and ELAN, designed with gradient path planning in mind. The Generalized Efficient Layer Aggregation Network (GELAN) enhances lightweight design, inference speed, and accuracy.

\subsection{Evaluation Metrics}
PPE detection models are evaluated using various metrics to measure their performance accurately. The Microsoft Common Objects in Context (MS-COCO) dataset employs several common metrics, including Precision (P) and Recall (R).

Additionally, Mean Average Precision (mAP) assesses the detection accuracy across all classes. It is calculated by determining Precision (P) and Recall (R) for each class and then averaging these values to provide an overall score. 
The Intersection over Union (IoU) metric is used to measure the accuracy of object localization. 
IoU calculates the overlap between the ground truth bounding boxes $(b_g)$ and the model’s predicted bounding boxes $(b_{pred})$ as follows:
$$\mbox{IoU} = \frac{\mbox{Area}(b_{pred}\cap b_g)}{\mbox{Area}(b_{pred}\cup b_g)}$$
Where $b_g$ represents the ground truth bounding box, and $p_{pred}$ denotes the bounding box predicted by the model, AP50 (Average Precision at 50\% IoU) is a metric used in object detection to evaluate the precision and recall of a model at a single Intersection over Union (IoU) threshold of 50\%. 
\begin{table}[h!]
\begin{tabular}{|l|c|} 
\hline
\textbf{Class}   & \multicolumn{1}{l|}{\textbf{Instances}}  \\ 
\hline
no-safety-glove  & 245                                      \\ 
\hline
no-safety-helmet & 2905                                     \\ 
\hline
no-safety-shoes  & 3341                                     \\ 
\hline
safety-glove     & 1973                                     \\ 
\hline
safety-helmet    & 5289                                     \\ 
\hline
safety-shoes     & 6066                                     \\ 
\hline
welding-helmet   & 135                                      \\
\hline
\end{tabular}
\caption{Class distribution for PPE detection}
\label{tab: ppe dist}
\end{table}

% \vspace{-0.5cm}
This means that a detected object's bounding box is considered a true positive if its IoU with the ground-truth bounding box is at least 50\%. AP50-95 evaluates the model's performance across multiple IoU thresholds. 
It averages the Average Precision (AP) scores calculated at ten different IoU thresholds: 50\%, 55\%, 60\%, 65\%, 70\%, 75\%, 80\%, 85\%, 90\%, and 95\%. This range of thresholds provides a broader view of the model's ability to detect objects with varying degrees of overlap, from relatively loose (50\%) to very strict (95\%). The AP50-95 score is the mean of these AP values.

For evaluation of the overall approach of PPE detection through Action Recognition, a clip-level metric is defined. In a clip of 15 seconds, even if 1 frame of the 15 clips has any PPE violation detected, then the entire clip is considered as a clip having a violation. 

Let \( N \) be the total number of frames in a clip (in this case 3). Let \( p_i \) be a boolean indicator for frame \( i \), where \( p_i = 1 \) if there is at least one person detected without the required PPE in frame \( i \), and \( p_i = 0 \) otherwise. Finally, \( V \) is a boolean indicator for the clip, where \( V = 1 \) if the clip is considered as having a violation, and \( V = 0 \) otherwise.

$
\mbox{The metric is defined as follows: V} = \begin{cases} 
1 & \text{if} \sum_{i=1}^{N} p_i \geq 1 \\
0 & \text{otherwise}
\end{cases}
$

This means that $V = 1$ (violation in the clip) if at least one frame \( i \) has \( p_i = 1 \) (indicating a detected PPE violation) and \( V = 0 \) (no violation in the clip) if all frames \( i \) have \( p_i = 0 \) (indicating no detected PPE violations).

\subsection{Proposed Approach for Action Recognition-Based PPE Detection}
To effectively monitor and ensure compliance with industrial safety protocols, we integrate our action recognition and PPE detection models into an integrated framework, presented in Algorithm~\ref{algo:1}. 
First, An action recognition model (Slow-Fast Network) takes in the input video feed (V) and trains it to identify and indicate the location of activities in the video. 

\begin{algorithm}[H]
\caption{Integrated Action Recognition and PPE Detection Framework}
\label{algo:1}
\resizebox{\textwidth}{!}{
\begin{minipage}{\textwidth}
\SetAlgoLined
\KwIn{Video footage \( V \)}
\KwOut{Detection of PPE violations and safety certification}

\SetKwFunction{FAR}{ActionRecognitionModel}
\SetKwFunction{FPPE}{PPEDetectionModel}
\SetKwFunction{FCompliance}{PPEComplianceCheck\_Train}
\SetKwFunction{FHybrid}{PPECheck\_Inference}

\SetKwProg{Fn}{Function}{:}{}

\Fn{\FCompliance{$\{F_{\text{first}}, F_{\text{middle}}, F_{\text{last}}\},\\ B_{\text{info}}, \text{PPE\_List}$}}{
    \For{each frame \( F_i \) in \(\{F_{\text{first}}, F_{\text{middle}}, F_{\text{last}}\}\)}{
        \For{each \( B_{\text{info},j} \)}{
            \( \text{required\_PPE} \gets \text{PPE\_List}[a_j] \)\;\\
            \For{each \( p \) in \text{required\_PPE}}{
                \If{\( p \) not in \( B_{\text{PPE}} \)}{Mark as violation\;}
                \Else{Certify safety compliance\;}
            }
        }
    }
}
\;

\Fn{\FHybrid{$V$}}{
    frames\_info \( \gets \) \FAR{$V$}\;\\
    frames \( \gets \{F_{\text{first}}, F_{\text{middle}}, F_{\text{last}}\} \)\;\\
    \For{frame in frames}{
        \FPPE{frame, frames\_info}\;
    }
    \FCompliance{frames, frames\_info, \text{PPE\_List}}\;
}
\;

\end{minipage}
}

\end{algorithm}
This information is stored in frames\_info. These locations are fed into PPEDetectionModel (YOLOv9), which was trained on a custom dataset to identify PPE. 
Afterwards, a new module PPEComplainceCheck\_Train has been designed to check PPE compliance. This model takes three frames F\_{first}, F\_{middle}, F\_{last} and for each frame, it will check for PPE in each B\_info generated from the Action Recognition model.
These PPEs are detected and then checked with a list of PPE requirements required\_PPE from the PPE\_list dictionary. 
If the PPE detected is not in required\_PPE, then it is marked as a violation or else, It is certified as safety complied. 
PPECheck\_Inference has been designed to take the video feed (\textbf{V}) as input to run the inference and generate the compliance information.

The YOLOv9 model determines whether the individual in the indicated area is wearing the appropriate PPE or not. 
If the user possesses the appropriate PPE for the activity, the system displays a message certifying their safety. 
If not, the system indicates that the individual is missing certain safety equipment. 
This strategy allows us to reduce the false positives by looking for only PPE designated for activity detected.
\section{Experiments and Results}
We conducted action recognition experiments using the SlowFast network on a real-life industry environment dataset, for which no previous datasets are available. We benchmarked our dataset against existing state-of-the-art models. All experiments were performed on a single RTX A6000 GPU 48GB VRAM and Intel® Xeon(R) Gold 5318Y CPU @ 2.10GHz × 96.

\subsection{Action Recognition}
For the SlowFast network, we utilized the implementation from the official repository  \cite{fan2020pyslowfast} and trained the model for 50 epochs with fine-tuned hyperparameters. The results are summarized in Table 6 and Table 7. We observe that Recall@Top k metrics are high.

\begin{table}[h]
\centering
\begin{tabular}{|l|r|}
\hline
Total Extracted Frames       & 7920   \\ \hline
Recall @ IoU=0.5             & 0.5959 \\ \hline
Precision @ IoU=0.5          & 0.6002 \\ \hline
Recall @ Top 3               & 0.9423 \\ \hline
Recall @ Top 5               & 0.9820 \\ \hline

\end{tabular}
\caption{Testing Metrics: Summary of Video Action Recognition Results}
\end{table}

\vspace{-0.7cm}

\begin{table}[h!]
\centering
\begin{tabular}{|>{\centering\arraybackslash}m{3cm}|>{\centering\arraybackslash}m{3cm}|}
\hline
\textbf{Action Class} & \textbf{AP@0.5IOU} \\
\hline
Crane Movement & 0.0121 \\
\hline
Observing & 0.1001 \\
\hline
Interacting & 0.0845 \\
\hline
Walking & 0.1340 \\
\hline
Carrying an Object & 0.1181 \\
\hline
Moving on a Bike & 0.0670 \\
\hline
Moving on a Cycle & 0.5317 \\
\hline
Lifting an Object & 0.1585 \\
\hline
Handing over an Object & 0.01902 \\
\hline
Interacting with a Machine & 0.0214 \\
\hline
Pushing or Pulling an Object & 0.0300 \\
\hline
Welding & 0.0354 \\
\hline
\textbf{Mean AP@0.5IOU} & \textbf{0.1093} \\
\hline
\end{tabular}
\caption{Testing mAP@0.5IOU : Performance By Category. AP@0.5IOU for various action classes}
\label{table:AP@0.5IOU}
\end{table}

% We implemented several baseline approaches for PPE detection.
% \vspace{-1.0cm}
\subsection{PPE Detection}
Although there are datasets available for PPE detection, we are not aware of any real-life industry environment datasets. 
We benchmark the dataset with existing state-of-the-art models. The dataset for PPE detection mentioned in table \ref{tab: ppe dist} is formatted in MS COCO style. 

We utilized the implementations of RetinaNet and FastRCNN from detectron-2 \cite{wu2019detectron2} and trained them on our dataset for 20,000 and 27,000 iterations, respectively, using the default hyperparameters. 
For the YOLOv9 model \cite{wang2024yolov9}, we used the official implementation and trained it for 120 epochs with default hyperparameters.

\begin{table}[h!]
\begin{tabular}{|l|c|c|c|} 
\hline
\multicolumn{1}{|c|}{\textbf{Classes}} & \begin{tabular}[c]{@{}c@{}}\textbf{Retina}\\\textbf{ Net}\end{tabular} & \begin{tabular}[c]{@{}c@{}}\textbf{FastRCNN}\\\textbf{ (R101)}\end{tabular} & \textbf{YoloV9}  \\ 
\hline
no-safety-glove                        & 56.9                                                                   & 66.1                                                                        & 67.1             \\ 
\hline
no-safety-helmet                       & 64.0                                                                   & 67.3                                                                        & 76.9             \\ 
\hline
no-safety-shoes                        & 49.4                                                                   & 54.4                                                                        & 66.4             \\ 
\hline
safety-glove                           & 37.1                                                                   & 39.9                                                                        & 51.5             \\ 
\hline
safety-helmet                          & 68.9                                                                   & 65.8                                                                        & 75.8             \\ 
\hline
safety-shoes                           & 56.2                                                                   & 58.5                                                                        & 70.8             \\ 
\hline
welding-helmet                         & 46.3                                                                   & 55.6                                                                        & 62.7             \\
\hline
\end{tabular}
\caption{Average Precision (AP50-95) score of various models on our dataset}
\label{tab: ppe metric}
\end{table}

% \vspace{-0.5cm}
Results on PPE detection are shown in the table \ref{tab: ppe metric} and \ref{tab:ppe yolov9}. The performance is similar to those obtained by training PPE-detection models on SH17~\cite{Ahmad2024SH17AD}. We observe that the precision for the classes \textit{no-safety-helmet} and \textit{safety-helmet} is significantly higher compared to other classes. This is attributed to the distinct appearance of the helmet and the fact that these two classes have the highest sample count in the dataset, as shown in Table \ref{tab: ppe dist}. The class \textit{safety-glove} has the lowest precision, while \textit{no-safety-glove} exhibits comparatively higher precision. This disparity may be due to the relatively lower sample counts for safety gloves and the model potentially confusing gloves as extensions of clothing.

\begin{table}[h!]
\begin{tabular}{|l|c|c|c|} 
\hline
\textbf{Classes} & \textbf{P} & \textbf{R} & \textbf{F1}  \\ 
\hline
no-safety-glove  & 95.7       & 88.0       & 91.7         \\ 
\hline
no-safety-helmet & 98.8       & 98.2       & 98.5         \\ 
\hline
no-safety-shoes  & 88.6       & 87.3       & 87.9         \\ 
\hline
safety-glove     & 80.2       & 80.7       & 80.4         \\ 
\hline
safety-helmet    & 98.9       & 98.9       & 98.9         \\ 
\hline
safety-shoes     & 85.7       & 88.7       & 87.2         \\ 
\hline
welding-helmet   & 93.2       & 92.0       & 92.6         \\
\hline
\end{tabular}
\caption{Precision, Recall and F1-Score of YOLOv9 models on the test dataset}
\label{tab:ppe yolov9}
\end{table}

% \vspace{-0.9cm}
\subsection{Action Recognition based PPE Detection}
To assess the effectiveness of PPE violation detection using the Action Recognition approach, we explored three different methods: (1) detecting common PPE items such as safety shoes and helmets, (2) detecting all types of PPE, and (3) leveraging Action Recognition for PPE detection with violation on at least 1 frame and 2 frames from the algorithm~\ref{algo:1}. The dataset with $109$ videos consisting of $54$ videos with violations is being used. These clips were chosen such that the field of view was sufficient enough to get a good focus on workers, and activities. We also ensured coverage of different types of actions and violations. The results of these approaches are presented in Table \ref{tab: 3approaches}.

\begin{figure}
    \centering
    \includegraphics[width=0.75\linewidth]{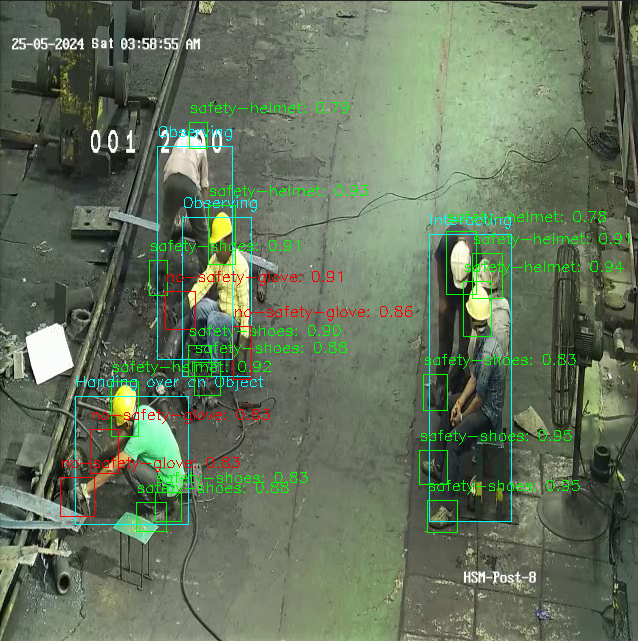}
    \caption{Combined action recognition and PPE detection for real-time safety compliance.}
    \label{fig:enter-label}
\end{figure}

\begin{table}[h!]
\centering
\begin{tabular}{|l|l|l|l|} 
\hline
\textbf{Algorithm Type}                                                                  & \textbf{Precision} & \textbf{Recall} & \textbf{F1 Score}  \\ 
\hline
\textbf{Common PPE}                                                                      & 0.62               & 0.55            & 0.59               \\ 
\hline
\textbf{All PPE}                                                                         & 0.51               & 0.56            & 0.54               \\ 
\hline
\begin{tabular}[c]{@{}l@{}}\textbf{Activity based PPE }\\\textbf{(1 frame)}\end{tabular} & 0.60               & \textbf{0.93}   & \textbf{0.73}      \\ 
\hline
\begin{tabular}[c]{@{}l@{}}\textbf{Activity based PPE}\\\textbf{~(2 frame)}\end{tabular} & \textbf{0.64}      & 0.83            & 0.72               \\
\hline
\end{tabular}
\caption{Precision, Recall and F1 Score on all three approaches.}
\label{tab: 3approaches}
\end{table}

While state-of-the-art models like YOLO are effective at detecting PPE, they tend to generate a high number of false positives. This is because not all PPE is required at all times; the necessity depends on the specific action being performed. For example, a person walking on a workshop floor does not need a welding helmet. We observe better precision when detecting common PPE (e.g., safety shoes, helmets) compared to detecting all PPE. This issue can be addressed through an action-based approach, which identifies the necessary PPE based on the activity being performed and checks only for those specific items. As shown in the table \ref{tab: 3approaches}, the activity-based approach achieves a high recall rate of $93\%$ by focusing on the relevant PPE for each action. By analyzing at least two frames to determine a violation, false positives are further reduced, leading to an increase in precision.

In an industrial environment, safety must be the top priority. Therefore, our primary goal is to maximize the detection of videos with violations. As a result, we focus on optimizing the recall metric. All violations detected by the algorithm would be verified by the safety officer before initiating disciplinary action.
% In the industry, missing violations can become costlier in the form of loss of human life and our approach tries to minimize the False Negatives and improve the recall. 

We also conducted a comparative study between our algorithm and human evaluators. 28 Human evaluators were given guidelines on how to detect violations and were asked to identify violations from a set of randomly selected 20 video clips (sample image Fig~\ref{fig:enter-label}). The ground truth of these samples is verified by experts from the safety department. Overall human evaluators' average precision of 1 (based on the majority answer) and an average recall of 0.78\% (based on the majority answer) indicate that human evaluators are able to accurately identify the True Positives (Violations being identified as violations) but missed some of the violations (Violations identified as Non-violations). Our proposed model gave a precision of 81.2\% and a recall of 93\%. 

\section{Discussion}
The inference runs with an average time of 117.25 ms per frame, and a total inference time of 1.76 seconds to process a 15-second video clip. During the process, CPU usage peaks at 2\%, memory usage is 5.6\%, and GPU usage peaks at 12\%. The CCTV cameras used in the study, record data at 12fps, and we have tested that the above system (see section 5) can process 25 video streams concurrently in real-time.

One of the limitations of our dataset is having only a 2D RGB video feed. This dataset lacks depth and 3D video feeds which can be used for a more comprehensive understanding of the working environment and context. For example, a person who is assisting the crane movement should not be directly below the crane and should maintain a certain distance from the path. This distance calculation between the worker and crane is difficult to measure accurately because camera positioning will impact the accuracy. We plan to deploy depth sensors at designated sites and collect the data for further augmenting the dataset.

\begin{comment}

%%%
\begin{table}[h!]
\centering
\begin{tabular}{|c|c|c|c|c|}
\hline
\textbf{V\_File} & \textbf{G\_Truth} & \textbf{HIL Result} & \textbf{Model Result} & \textbf{Avg. Error\%} \\
\hline
1.mp4  & V  & V  & V  & 13.79 \\
2.mp4  & V  & V  & V  & 17.24 \\
3.mp4  & V  & V  & V  & 10.34 \\
4.mp4  & V  & V  & V  & 24.14 \\
5.mp4  & V  & V  & V  & 51.72 \\
6.mp4  & V  & V  & V  & 27.59 \\
7.mp4  & NV & NV & V  & 17.24 \\
8.mp4  & NV & NV & V  & 10.34 \\
9.mp4  & V  & V  & V  & 24.14 \\
10.mp4 & V  & NV & V  & 24.14 \\
11.mp4 & NV & NV & NV & 51.72 \\
12.mp4 & V  & V  & NV & 41.38 \\
13.mp4 & NV & NV & V  & 24.14 \\
14.mp4 & NV & NV & V  & 20.69 \\
15.mp4 & V  & V  & V  & 20.69 \\
16.mp4 & V  & V  & V  & 20.69 \\
17.mp4 & NV & NV & V  & 17.24 \\
18.mp4 & V  & NV & NV & 44.83 \\
19.mp4 & V  & V  & V  & 20.69 \\
20.mp4 & V  & NV & NV & 48.28 \\
\hline
\end{tabular}
%%%%%

\caption{Human In Loop vs. Model Study Results of 20 Random Videos. G\_Truth are the ground truth labels given by Safety experts, HIL Result refers to the majority labels given by the humans in the loop. V stands for Violation and NV stands for Non-violation label at a clip level. Avg. Error is the \% of humans that made an error for that clip.}
\label{table:hil_results}
\end{table}
\end{comment}

The other issue is the field of view (FOV). The usual camera setup in the industrial environment is to cover the maximum area possible. 
Due to this, the field of view becomes large, and small objects such as gloves, glasses, and shoes for the distant objects are difficult to detect. 
One solution to solve this is to increase the number of cameras. 
From analyzing the videos, we have observed that a FOV of 20mts is sufficient enough to get good accuracy. 
Practically also, this approach makes sense as many large plants have shop floors covered in huge areas and complete coverage might not be feasible financially.

\section{Conclusion}

In this paper, we present a novel dataset with dense spatio-temporal annotations designed to recognize industrial actions. This dataset sets itself apart from existing action detection datasets by offering a diverse and realistic collection of industrial environment clips, as well as comprehensive annotations for commonly performed industrial tasks. We also introduce an innovative approach for detecting PPE violations by integrating action recognition with object detection models. Our approach achieves high recall at the clip level. Our work highlights the need for further research on Industrial Action Recognition (IAR) and aims to inspire continued exploration in application areas such as human workflow analysis and human-machine interactions.

\begin{acks}
The authors thank SAIL, Bokaro Steel Plant, and Deepnet Analytics LLP for their support in data set preparation. The authors also thank Himanshu from IIT Bhilai for his contributions in experimentation. We are also grateful to all of the reviewers for their valuable comments and suggestions to improve the scientific value of the paper.
\end{acks}

\bibliographystyle{ACM-Reference-Format}
\bibliography{references}
\end{document}